\documentclass[11pt]{article}
\usepackage[utf8]{inputenc}
\usepackage[T1]{fontenc}
\usepackage{times}
\usepackage[margin=0.78in]{geometry}
\usepackage{graphicx}
\usepackage{amsmath,amssymb}
\usepackage{booktabs}
\usepackage{tabularx}
\usepackage[numbers]{natbib}
\usepackage[hidelinks]{hyperref}
\usepackage{float}
\usepackage{xcolor}
\usepackage{algorithm}
\usepackage{algpseudocode}

\title{\textbf{Governing Equations of Human Sleep Macro-Structure:\\
Symbolic Discovery from 1{,}106 Nights}}
\author{Junhui Gao\\[2ex]
\href{mailto:jhgao68@163.com}{jhgao68@163.com}\\[2ex]
Shenzhen Peng Brain Technology Co., Ltd.}
\date{}

\begin{document}
\maketitle

\begin{abstract}
Human sleep is organized into a small number of cycles per night: deep sleep
dominates early cycles, REM lengthens toward morning, and light sleep
clusters at cycle boundaries. These regularities have long been described
qualitatively but never distilled into closed-form laws from large-scale
data. A total of 1{,}106 nights (Sleep-EDF 153 + SHHS 953) are segmented
into REM-anchored cycles, and four cycle-indexed statistics follow simple
closed-form equations: cycle length declines nearly linearly (R$^2$=0.996),
REM duration follows an inverted U (R$^2$=0.873), deep-sleep (N3) run length
decays exponentially (R$^2$=0.987) with a time constant of 2.94 cycles,
consistent with the homeostatic Process S of the two-process model, and
light-sleep (N1) run length follows a U shape (R$^2$=0.956), confirming a
``long--short--long'' pattern across the night. The laws are reproduced by
per-night differentiable fitting (142/142 and 773/773 nights converge), by
unconstrained symbolic regression (PySR), and by a 500-night generative
synthesizer built from them. Their robustness is established against the
two main methodological threats: bootstrap CIs computed by resampling
whole nights (1{,}000 replications) exclude zero for every shape term, and
a five-level scan of the REM-merge segmentation threshold preserves every
qualitative form. These appear to be the first closed-form laws governing
the macro-structure of sleep, distilled and tested across an age-mixed
sample (healthy young adults and community-based older adults).
\end{abstract}

\section{Introduction}
\label{sec:intro}

Sleep is organized at two scales. At the micro-scale, the night is divided
into 30-second epochs and each epoch is assigned one of five stages (Wake,
N1, N2, N3, REM) by standardized scoring rules \citep{rechtschaffen1968,iber2007aasm}. This
epoch-level sequence is what machine learning has learned to predict with
increasing accuracy, from single-channel CNNs \citep{supratak2017deepsleepnet}
to fully convolutional and transformer architectures
\citep{perslev2019utime,perslev2021usleep,phan2022sleeptransformer}. At the macro-scale, the
same night is organized into 4--6 sleep cycles, each containing a progression
from light sleep through deep sleep to REM, and the properties of these
cycles change systematically across the night. Deep-sleep (slow-wave)
intensity is maximal in the first cycle and declines toward morning; REM
episodes lengthen across the night; and light sleep (N1) punctuates the
transitions between deeper stages and REM \citep{dement1957,carskadon2005,mccarley1975}.

This macro-scale organization is not merely descriptive folklore. It is the
core prediction surface of the two-process model of sleep regulation
\citep{borbely1982,borbely2016two}, in which a homeostatic process
(Process S) accumulates during wakefulness and decays during sleep, and a
circadian process (Process C) modulates sleep propensity
\citep{dijk1995}. The model
successfully explains the timing of slow-wave activity and REM
\citep{achermann2004,borbely2016two}, and quantitative sleep-regulation
models instantiate its assumptions as coupled differential equations
\citep{achermann2004}. Yet the gap addressed here is that the
\emph{structure-level} regularities of the hypnogram itself (how long
each stage persists in the $k$-th cycle of the night) have never been
extracted as closed-form laws from large-scale polysomnography. The curves
visibly decay, peak, or rise; no closed-form expression has
been written down and tested across thousands of nights.

Two consequences follow from the absence of such laws. First, the
understanding of sleep architecture remains qualitative: statements such as
``deep sleep is front-loaded'' cannot be falsified, compared across
datasets, or linked to model parameters. Second, the
discovery-style questions of sleep science (is the N3 decline exponential?
does N1 follow a U shape?) remain unquantified, even though the data to
answer them have existed in public repositories for two decades.

The pipeline developed here addresses these gaps by applying the AI-for-science paradigm
of distilling closed-form empirical laws from data
\citep{schmidt2009distilling,brunton2016sindy,udrescu2020aifeynman,cranmer2020pysr}
to the hypnogram itself. This work:

\begin{enumerate}
\item \textbf{Segment} 1{,}106 nights (Sleep-EDF Expanded 153 nights,
SHHS 953 nights) into REM-anchored sleep cycles with a transparent,
reproducible algorithm, and define four cycle-indexed macro-structure
statistics: cycle length, total REM duration, longest N3 run, and mean N1
run length.
\item \textbf{Discover} closed-form empirical laws for all four
statistics by two complementary levels of library fitting (population
means and per-night differentiable fits) plus an unconstrained symbolic
search (PySR) that imposes no template, and validate the
laws by generating 500 synthetic nights from them.
\end{enumerate}

Throughout, \emph{governing} is used descriptively: the equations summarize
the per-cycle evolution of stage-run statistics and are empirical fits
rather than mechanistic derivations.

The main results are four closed-form laws (Section~\ref{sec:r-eq}): cycle
length declines with a slight, near-linear downward curvature (R$^2$=0.996), REM follows an inverted U
(R$^2$=0.873), N3 runs decay exponentially with a time constant of 2.94
cycles (R$^2$=0.987), and N1 run length follows a U shape (R$^2$=0.956),
quantitatively confirming the ``long--short--long'' N1 pattern across the
night. The laws are verified individually per night, reproduced without a
template by PySR, and support a 500-night synthesizer whose statistics match
real data. Bootstrap CIs computed by resampling whole nights exclude zero
for every shape term, and a five-level scan of the segmentation threshold
preserves all four qualitative forms, so the shapes are not artifacts of
the extraction rule.

\section{Related work}
\label{sec:related}

\paragraph{Sleep-cycle structure has been characterized descriptively.}
Quantification of sleep macro-structure goes back to Feinberg's 1974
demonstration that sleep-cycle patterns change with age and that within-night
cycle composition is highly organized \citep{feinberg1974};
Carskadon and Dement's textbook chapter fixed the canonical description of
normal human sleep architecture (front-loaded deep sleep, lengthening REM,
4--6 cycles per night) \citep{carskadon2005}; and forced-desynchrony
experiments separated circadian and homeostatic contributions to sleep
structure, including slow-wave and spindle activity \citep{dijk1995}. More
recently, data-driven measures have re-parameterized REM propensity and
inter-REM intervals across species \citep{akhavan2026rems}. What these
bodies of work share is description and mechanism at the level of
aggregates or parameters; none has reduced the \emph{per-cycle evolution of
stage-run statistics} to closed-form laws fitted and tested on
thousands of nights. The contribution is that distillation step.

\paragraph{Template-free symbolic regression.}
Discovering closed-form laws from data has a mature toolkit: nonlinear
system identification (SINDy) \citep{brunton2016sindy}, free-form law
distillation \citep{schmidt2009distilling}, physics-inspired symbolic
regression \citep{udrescu2020aifeynman}, and evolutionary symbolic search
(PySR) \citep{cranmer2020pysr}. In sleep science the quantitative tradition
has mostly built mechanistic models by hand (e.g., the two-process model
\citep{borbely1982,borbely2016two,achermann2004}) rather than fitting
closed forms directly to macro-structural statistics. The pipeline applies
the AI-for-science loop to the hypnogram itself.

\section{Methods}
\label{sec:methods}

\begin{figure}[htbp]
\centering
\includegraphics[width=0.62\linewidth]{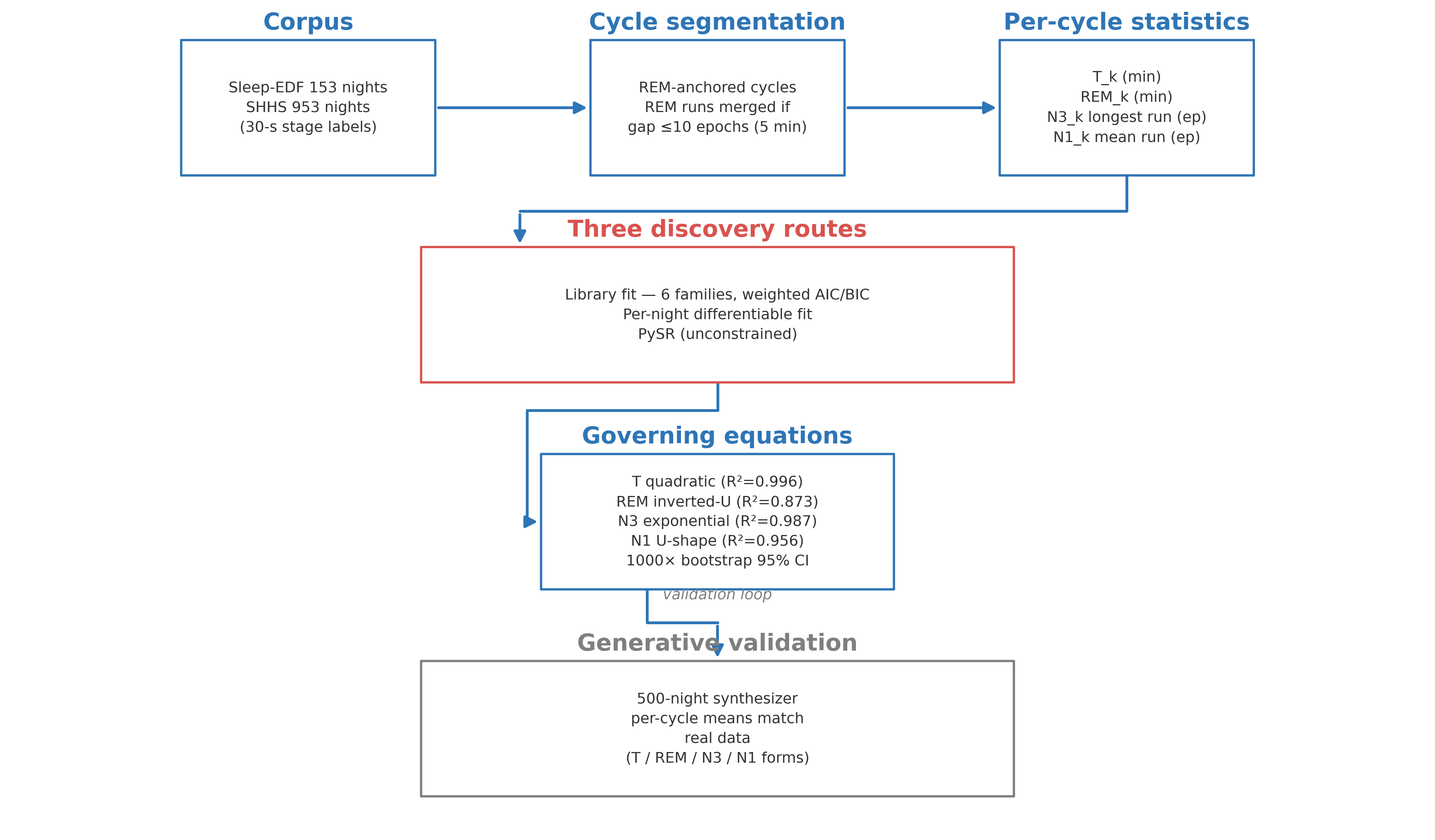}
\caption{Analysis pipeline. Stage labels from Sleep-EDF and SHHS are
segmented into REM-anchored cycles; four cycle-indexed statistics are
aggregated as per-cycle means, and two levels of library fitting
(population means and per-night fits) plus an unconstrained PySR search
converge on closed-form governing equations with bootstrap 95\% CIs. The
laws are validated by a 500-night synthesizer.}
\label{fig:pipeline}
\end{figure}

The pipeline is summarised in Fig.~\ref{fig:pipeline}: stage labels are
segmented into REM-anchored cycles, four cycle-indexed statistics are
fitted by two library levels plus an unconstrained search, and the resulting
laws are validated by generation.

\subsection{Data and preprocessing}
\label{sec:data}

\textbf{Sleep-EDF Expanded (SEDF).} The analysis uses the 153 SC-subject nights of the
Sleep-EDF Expanded database \citep{kemp2000edf}, with
single-channel FPz-Cz EEG
sampled at 100~Hz, scored in 30-second epochs by the Rechtschaffen--Kales
rules into Wake, N1, N2, N3, REM, and movement/unknown categories, keeping
only epochs with a valid stage in $\{0,\dots,4\}$ (Wake$\dots$REM).
The 153 nights produce 704 cycles after segmentation (Section~\ref{sec:cycles}).

\textbf{SHHS.} The Sleep Heart Health Study \citep{quan1997shhs} provides
overnight polysomnography with C4-A1 EEG at 125~Hz. From this cohort the pipeline takes 953
nights, downsamples EEG to 100~Hz (polyphase resampling,
factor 4/5), and maps the SHHS stage codes to the same five-stage
scheme, again discarding invalid epochs. Combined, the corpus covers
1{,}106 nights and $>$1.5M epochs.

Both pipelines operate on labels rather than raw signal quality: the
structural statistics are label-level and therefore insensitive to amplifier
noise, movement artifacts, and montage differences. No EEG cleaning beyond
stage filtering is applied.

\subsection{Cycle segmentation}
\label{sec:cycles}

\begin{figure}[H]
\centering
\includegraphics[width=0.62\linewidth]{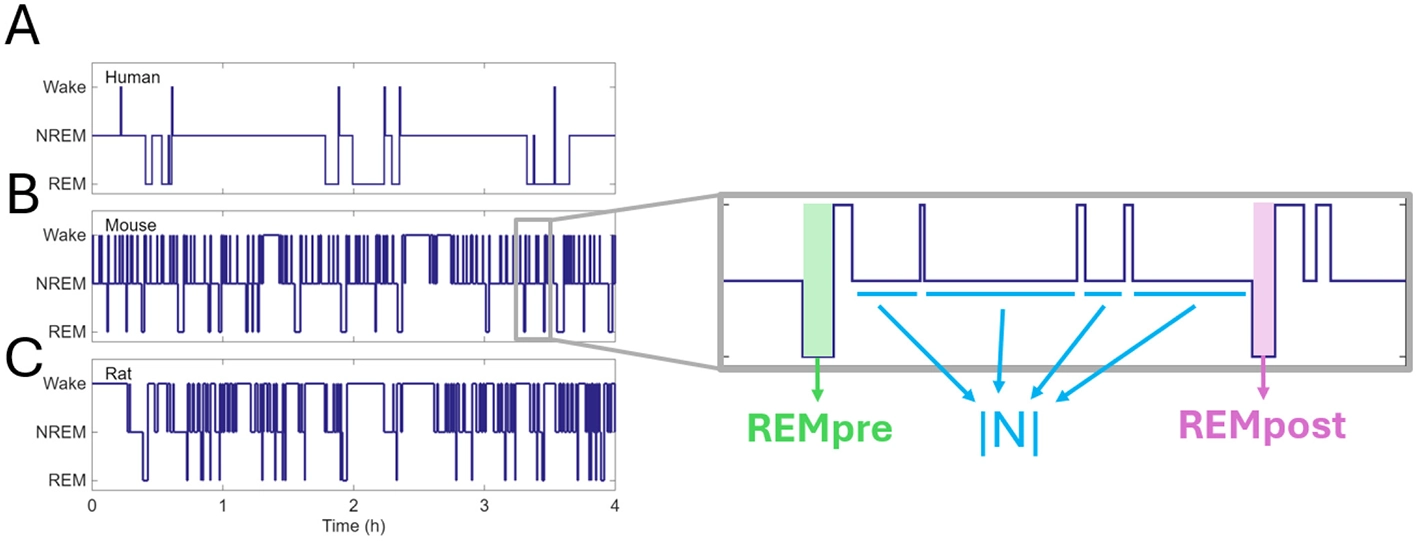}
\caption{Concept of the sleep cycle as a REM-to-REM interval, reproduced
from \citeauthor{akhavan2026rems} (\citeyear{akhavan2026rems}) under
CC BY~4.0. Hypnograms over 4\,h for human (A), mouse (B) and rat (C);
the inset defines a REM-anchored cycle as the interval from the onset of a
REM episode (REMpre, green) to the onset of the next REM episode
(REMpost, magenta), with the intervening NREM accumulation (INI, blue) the
putative driver of REM propensity. The cycle segmentation
(Sec.~\ref{sec:cycles}) shares this REM-to-REM anchoring; Sleep-EDF and
SHHS nights are scored at 30-s epochs with stages W/N1/N2/N3/REM.}
\label{fig:hypno}
\end{figure}

A sleep cycle is defined as a REM-to-REM interval
(Fig.~\ref{fig:hypno}). Raw hypnograms contain
spurious short interruptions of REM and adjacent stages, so a naive
``between two REM runs'' definition produces unstable cycle boundaries. Cycle
extraction therefore proceeds in two steps.

\paragraph{Step 1: REM-run consolidation.} The night is first split into
runs of consecutive identical stages. Two consecutive REM runs are merged when
the gap between them is at most 10 epochs (5 min), regardless of which
non-REM stages intervene; merged REM periods containing fewer than 2 epochs
of total REM are discarded. This removes micro-arousal interruptions while
preserving genuine REM--NREM alternations. Section~\ref{sec:r-eq} probes the
sensitivity of the inferred laws to this threshold.

\begin{figure}[H]
\centering
\includegraphics[width=0.62\linewidth]{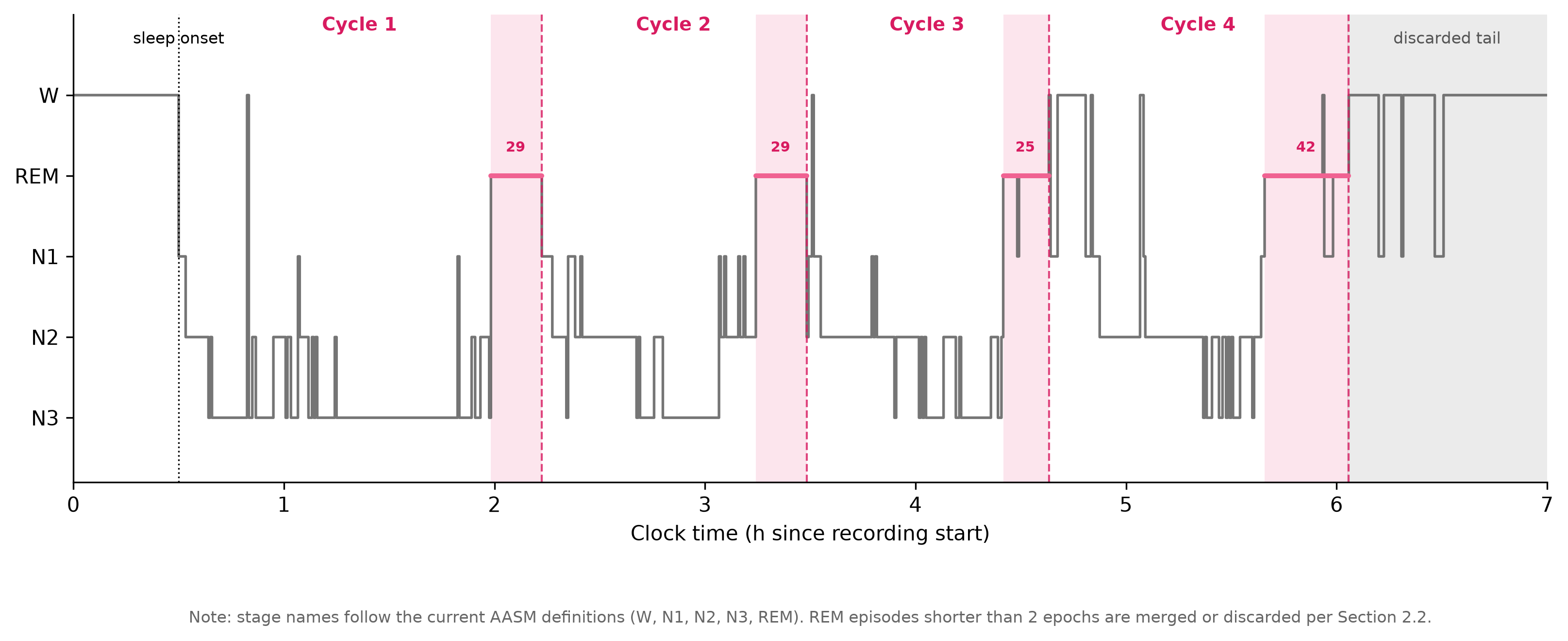}
\caption{Example of the proposed cycle segmentation on a real night
(Sleep-EDF, SC4001; line-style hypnogram, y-axis W/REM/N1/N2/N3 top to
bottom, REM shaded pink). Four cycles are delimited by dashed vertical lines at each
consolidated REM-run end; the final segment after the last REM run (gray)
is discarded because it has no terminating REM boundary. Run lengths of
each REM episode are labeled in epochs.}
\label{fig:segmented}
\end{figure}

\paragraph{Step 2: Cycle extraction.} Each consolidated REM run defines the
end of a cycle (Fig.~\ref{fig:segmented}). The $k$-th cycle spans epochs
$[\text{end of previous REM}+1,\ \text{end of current REM}]$; the $k$-th
cycle's index is $k=1,\dots,K$. The incomplete tail after the final REM run
(the final descending segment toward morning wake) is discarded, as it has
no terminating REM boundary. This gives $K$ between 1 and 9 (median 4;
$\approx$80\% of nights in 3--6) and 4{,}314 cycles in total.

The algorithm is deterministic, has no free parameters tuned on either
dataset, and is applied identically to SEDF and SHHS. The same code is
released.

\begin{algorithm}[H]
\caption{Cycle segmentation of a scored night}
\label{alg:segmentation}
\begin{algorithmic}[1]
\Require scored 30-s epoch sequence $s_1,\dots,s_N$ with stages W/N1/N2/N3/REM
\Ensure cycles $C_1,\dots,C_K$ and per-cycle statistics $T_k$, $\mathrm{REM}_k$, $\mathrm{N3}_k$, $\mathrm{N1}_k$
\State split the night into maximal runs of consecutive identical stages
\For{each pair of consecutive REM runs $r_i, r_{i+1}$}
  \If{the gap $\mathrm{end}(r_i)+1,\dots,\mathrm{start}(r_{i+1})-1$ spans at most 10 epochs}
    \State merge $r_i$ and $r_{i+1}$ into one consolidated REM run
  \EndIf
\EndFor
\State discard consolidated REM runs whose total REM is fewer than 2 epochs
\State let $R_1,\dots,R_K$ be the surviving REM runs in chronological order
\For{$k = 1 \dots K$}
  \If{$k = 1$}
    \State $C_1 \gets$ epochs from night start to $\mathrm{end}(R_1)$
  \Else
    \State $C_k \gets$ epochs from $\mathrm{end}(R_{k-1})+1$ to $\mathrm{end}(R_k)$
  \EndIf
\EndFor
\State discard the trailing epochs after $\mathrm{end}(R_K)$ (no terminating REM boundary)
\For{each cycle $C_k$}
  \State compute $T_k$, $\mathrm{REM}_k$, $\mathrm{N3}_k$, $\mathrm{N1}_k$ (Sec.~\ref{sec:indices})
\EndFor
\end{algorithmic}
\end{algorithm}

\subsection{Four macro-structure statistics}
\label{sec:indices}

For each cycle $k$ of a night, four statistics are computed:
\begin{itemize}
\item $T_k$: cycle length in minutes (epochs $\times$ 0.5);
\item $\text{REM}_k$: total REM duration in minutes within the cycle;
\item $\text{N3}_k$: length in epochs of the longest N3 run in the cycle;
\item $\text{N1}_k$: mean length in epochs of N1 runs in the cycle (the
  average run length, not the total amount, so that it measures
  \emph{persistence} of the transitional stage).
\end{itemize}
Per-cycle means across nights define population trajectories
$\mu_X(k)$ for $X\in\{T,\text{REM},\text{N3},\text{N1}\}$. The primary
hypothesis on N1 is a U shape (``long--short--long''): $\mu_{\text{N1}}$
decreases from cycle 1, reaches a trough mid-night, and recovers toward the
morning. Fitting a quadratic in $k$ and comparing it against a constant and
a linear model with AIC/BIC tests this directly (Section~\ref{sec:sr}).

\subsection{Symbolic regression: two levels plus an unconstrained search}
\label{sec:sr}

The candidate laws are fitted at two levels (pooled population means and
individual nights) and the discovery is then re-run without any template.
The
two library-based levels are complementary views of the same function
library rather than independent searches; the unconstrained PySR pass is the
genuinely independent route.

\textbf{Population-level library fit.} A small library of
closed forms is fitted to the pooled per-cycle means with cycle index
$x=k-1$:
linear $p_0+p_1 x$; quadratic $p_0+p_1 x+p_2 x^2$; exponential
$c+a e^{-b x}$; reverse exponential $c-a e^{-b x}$; power
$c+a(x+1)^{-b}$; and Michaelis--Menten-like $c+\frac{ax}{b+x}$. Fits are
weighted by the number of nights contributing to each cycle index, and model
selection uses AIC and BIC (effective sample size $n_{\text{eff}}=\sum
\text{weights}$). The fits cover SEDF, SHHS, and the pooled
corpus. For N3, a quadratic and an exponential both fit well; the
exponential is reported (see Section~\ref{sec:r-eq} for the choice
rationale and Fig.~\ref{fig:gov}).

\textbf{Per-night differentiable fit.} To test whether the laws
hold within single subjects, every night receives an individual fit. For a
night with cycles $k=1,\dots,K$, the scipy curve\_fit routine minimizes
the squared error of the four statistics against the candidate forms with
physiological initial values, recording convergence and fitted parameters.
This produces per-night parameter distributions (Fig.~\ref{fig:param}) and
tests cross-dataset parameter stability (Section~\ref{sec:r-indiv}).

\textbf{Unconstrained symbolic regression (PySR).} To rule out
that the laws are artifacts of the assumed library, an unconstrained PySR
run repeats the search \citep{cranmer2020pysr} on the pooled per-cycle
means with a general operator library ($+,-,\times,\div,\exp$, no
template), 40 populations $\times$ 20 generations, and checks whether the
discovered expressions coincide with the library forms
(Fig.~\ref{fig:pysr}).

\subsection{Generative validation: a 500-night synthesizer}
\label{sec:synth}

A simulator lays out complete night-long stage sequences from the fitted
laws. For each synthetic night it (i) draws the number of cycles from the
empirical distribution, (ii) draws per-cycle $T_k$, $\text{REM}_k$,
$\text{N3}_k$, $\text{N1}_k$ from the law-predicted means with lognormal
noise whose scale is fitted to the real per-cycle dispersion, (iii) converts
these into run lengths (placing N3 runs at the cycle front, REM at the
cycle end, N1 runs at the stage boundaries, and N2/Wake filling the
remainder), and (iv) emits the 30-second epoch sequence. The simulator
generates 500 nights and compares their per-cycle statistics against real
SHHS and SEDF (Fig.~\ref{fig:synth}).

\section{Results}
\label{sec:results}

\subsection{Cycle structure of 1{,}106 nights}
\label{sec:r-cycle}

Across the corpus the segmentation extracts 4{,}314 cycles (median 4 per
night; $\approx$80\% of nights in 3--6). Individual nights deviate
considerably from these medians (K spans 1 to 9), so the curves below
describe the population trajectory, not a universal template. Cycle length
decays from $\approx$126 min
(cycle 1, pooled) to $\approx$26 min (cycle 9), with the first cycle
typically lasting 2--3$\times$ longer than the last; total REM peaks early
at $\approx$19.9 min around cycles 2--3 before declining to $\approx$8.6
min by cycle 7; the longest N3 run falls from 18.5 epochs (Sleep-EDF cycle
1) / 26.9 epochs (SHHS) to essentially zero by cycle 9. These raw
trajectories (Fig.~\ref{fig:gov}, points) reproduce the classic arch of
sleep architecture \citep{borbely2016two} and establish the
population-level pattern quantified below.

\subsection{N1 run length: the ``long--short--long'' U shape}
\label{sec:r-n1}

\begin{figure}[!htbp]
\centering
\includegraphics[width=0.50\linewidth]{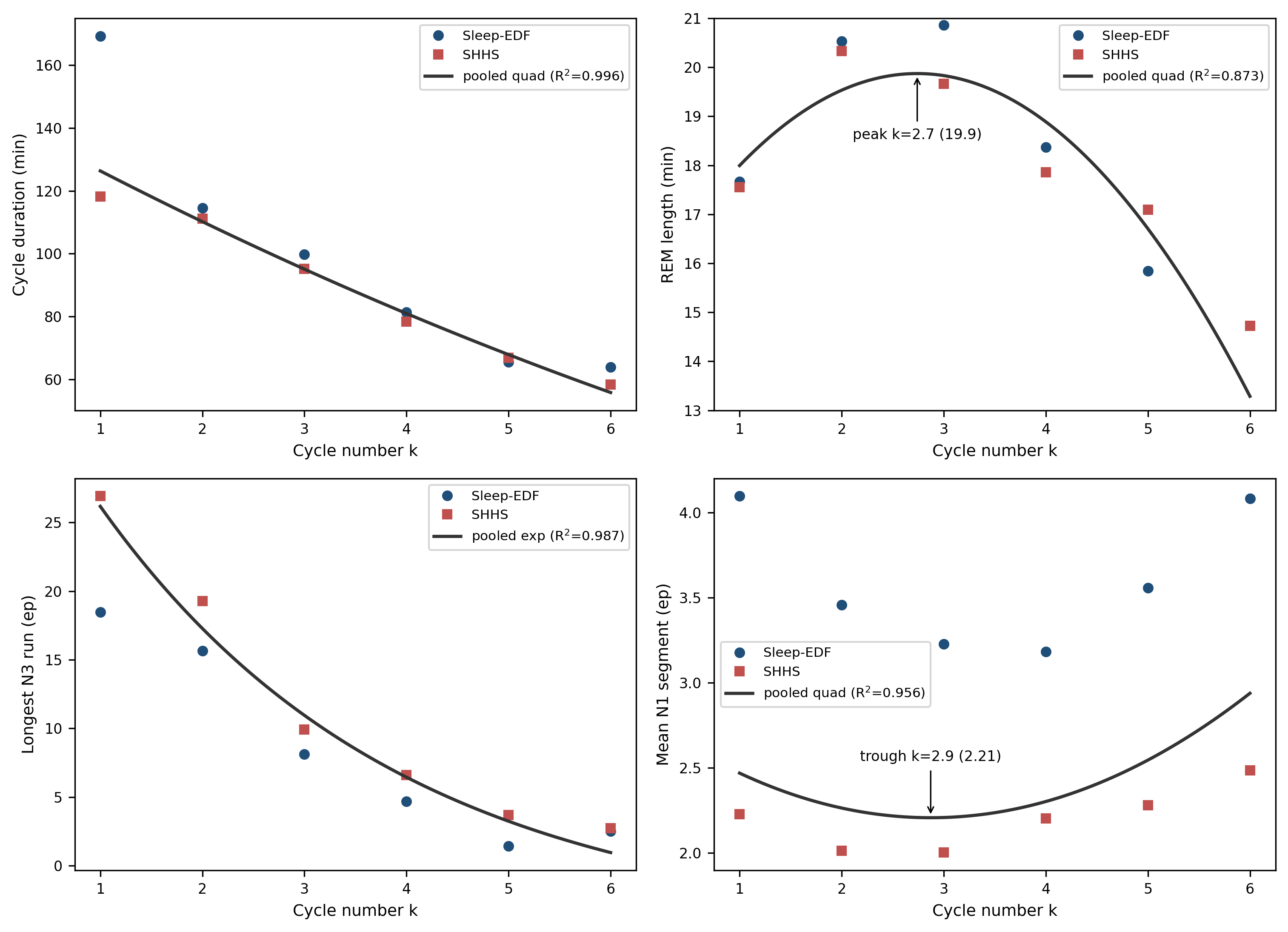}
\caption{Governing equations of sleep macro-structure. Points: per-cycle
means across Sleep-EDF ($n$=153 nights, circles) and SHHS ($n$=953 nights,
squares); curves: pooled weighted best-fit (Table~\ref{tab:eq}); N3 is shown
as the exponential form. R$^2$ as annotated: $T$ 0.996, REM 0.873, N3 0.987,
N1 0.956.}
\label{fig:gov}
\end{figure}

The mean N1 run length across cycle index is $4.10 \to 3.18 \to 4.20$
epochs (Sleep-EDF, cycles 1 / 4 / 7) and $2.23 \to 2.37$ (SHHS, cycles 1
vs 7): a trough in mid-night and recovery toward the morning. N1 occupies
only a few percent of scored epochs, so these means aggregate short,
fragmented runs; the U shape nevertheless appears in both cohorts. A
quadratic model fits the pooled trajectory with R$^2$=0.956 and beats the linear
model by $\approx$15 AIC points (the linear fit attains R$^2$=0.26). The
U shape therefore holds in two independent cohorts and at the pooled
level, more shallowly in SHHS, confirming the ``long--short--long''
pattern quantitatively at the population scale (Fig.~\ref{fig:gov},
bottom-right panel). This is consistent with N1\textquoteright s transitional
role: long runs buffer entry into deep sleep early in the night, shrink as
REM--NREM alternation stabilizes mid-night, and reappear around the
longer, more fragmented transitions of late cycles.

\subsection{Governing equations}
\label{sec:r-eq}

Table~\ref{tab:eq} lists the pooled closed-form laws ($x=k-1$).

\begin{table}[!htbp]
\centering
\caption{Closed-form governing equations of sleep macro-structure (pooled,
1{,}106 nights; $x=k-1$). $T$ and REM in minutes, N3 and N1 in epochs.
Weighted AIC/BIC selection over six families (linear, quadratic,
exponential, reverse exponential, power, Michaelis--Menten).}
\label{tab:eq}
\small
\begin{tabular}{l >{\centering\arraybackslash}p{2.0in} ccc}
\toprule
Statistic & Equation & R$^2$ & AIC & BIC \\
\midrule
Cycle length $T$ & $126.34 - 16.64 x + 0.509 x^2$ & 0.996 & 10.75 & 10.13 \\
REM duration & $17.99 + 2.158 x - 0.620 x^2$ & 0.873 & 0.39 & --0.24 \\
N3 longest run & $30.85\, e^{-0.340 x} - 4.68$ & 0.987 & 6.41 & 5.79 \\
N1 run length & $2.469 - 0.280 x + 0.075 x^2$ & 0.956 & --32.73 & --33.35 \\
\bottomrule
\end{tabular}
\smallskip
\begin{flushleft}
\footnotesize AIC/BIC are weighted (effective $n_{\text{eff}}=\sum$ weights;
lower is better) and compare models of the \emph{same} statistic only; they
are not comparable across rows. For N3 the quadratic alternative attains
AIC 4.13 / BIC 3.51 (R$^2$=0.991), a $\Delta$BIC of +2.3 against the
exponential; the fit selects the exponential for its physiological interpretation
(Process-S-like decay), its match to per-night fits, and its rediscovery by
PySR. Selection against the linear alternative: $\Delta$BIC $=8.3$ (REM),
$\ge$7.4 (N3) and $\Delta$AIC $\approx$15 (N1); for cycle length the
quadratic vs linear $\Delta$BIC $=0.7$, so the curvature term (whose CI
includes zero) is reported but flagged as near-linear.
\end{flushleft}
\end{table}

\begin{itemize}
\item \textbf{Cycle length} is almost linear with a slight downward
curvature: $T=126.34-16.64x+0.509x^2$ (R$^2$=0.996; 95\% bootstrap CIs:
$a$ [122.2, 130.5], $b$ [--20.5, --12.9], $c$ [--0.23, 1.26]). The linear
coefficient is consistently negative, but the curvature CI includes zero and
the BIC gain over a pure line is only 0.7 (R$^2$ 0.996 vs 0.994), so the
decline is statistically near-linear; the 16.6 min-per-cycle decrement
means the first cycle lasts more than twice as long as the last.
\item \textbf{REM duration} follows an inverted U:
$\text{REM}=17.99+2.158x-0.620x^2$ (R$^2$=0.873; CIs: $a$ [17.4, 18.6],
$b$ [1.46, 2.92], $c$ [--0.78, --0.46]), the noisiest of the four fits,
with a vertex at $k\approx 2.7$:
REM lengthens early, then shortens toward morning, the mirror image of
the N3 decline. The vertex near the middle third of the night is in phase
with the circadian drive for REM propensity \citep{borbely2016two}. The
negative curvature is clear (CI excludes 0) and
the quadratic beats the linear fit by 8.3 BIC points.
\item \textbf{N3 runs} decay exponentially:
$\text{N3}=30.85\,e^{-0.340x}-4.68$ (R$^2$=0.987; CIs: $a$ [28.0, 34.6],
$b$ [0.27, 0.42], $c$ [--8.6, --1.7]). The time constant $1/b = 2.94$
cycles (95\% CI 2.4--3.7) falls in the same range as the two-process model's Process S, which decays
with a time constant of approximately 2.5--3 cycles
\citep{borbely2016two,achermann2004}, so the exponential decay of deep-sleep
run length is quantitatively consistent with the homeostatic hourglass.
\item \textbf{N1 run length} follows a U shape:
$\text{N1}=2.469-0.280x+0.075x^2$ (R$^2$=0.956; CIs: $a$ [2.39, 2.57],
$b$ [--0.39, --0.19], $c$ [0.052, 0.099]; the positive curvature is
clear (CI excludes 0)), confirming
Section~\ref{sec:r-n1} at the equation level.
\end{itemize}
Model selection (Table~\ref{tab:eq}) strongly prefers the reported forms
over linear alternatives for REM, N3 and N1 ($\Delta$BIC $\ge$ 7.4 vs
linear; $\Delta$AIC $\approx$ 15 for N1); for cycle length the quadratic
and linear fits are nearly indistinguishable (BIC $\Delta=0.7$). For N3 a
quadratic also fits (R$^2$=0.991, $\Delta$BIC $=+2.3$ for the exponential),
but the fit selects the exponential: it is physiologically
interpretable (a Process-S-like decay), matches the per-night decay-rate
family (Section~\ref{sec:r-indiv}), and is the form rediscovered by PySR
(Section~\ref{sec:r-pysr}).

\paragraph{Bootstrap uncertainty and segmentation sensitivity.}
Nights are resampled with replacement (1{,}000 replications; pooled
Sleep-EDF+SHHS, weighted as above) to obtain 95\% CIs for every
coefficient (reported above). All key shape terms (negative REM
curvature, positive N3 decay rate, positive N1 curvature) have CIs
excluding zero; the cycle-length curvature CI includes zero, matching
the near-linear reading of $T$.

The segmentation rule is the main free parameter of the extraction
pipeline, so its influence is probed directly. Re-segmenting both
cohorts at REM-merge gaps of 2, 5, 10, 20 and 30 epochs changes the
number of extracted cycles (5{,}748, 4{,}846, 4{,}314, 4{,}011 and
3{,}909) and the fitted coefficients, but preserves every qualitative
shape. The REM quadratic is the BIC-selected form at all five gaps with
negative curvature ($c$ from $-0.15$ to $-1.12$; vertex at cycle
$k\approx1.4$ to $2.9$). The N1 quadratic retains positive curvature at
gaps of 5 epochs and above ($c$ from 0.057 to 0.123); at gap 2 a
monotone half-rise (Michaelis--Menten) form is preferred instead, so the
N1 arch is reported for the primary 10-epoch rule. The N3 decay rate
$b$ rises monotonically as the gap tightens: 0.19, 0.23, 0.34, 0.51 and
0.69 for gaps 30, 20, 10, 5 and 2, respectively, with the exponential
direction unchanged. The cycle-length trend stays monotone and
near-linear under every rule (the quadratic reaches R$^2\ge$0.97 at all
gaps, although BIC prefers power or exponential forms at some gaps).
The exact coefficients therefore depend on the segmentation rule, while
the reported shapes are robust to it; the 10-epoch (5-min) gap serves as
the physiologically motivated default.

\subsection{Slow-wave sleep N3: an exponential law tied to the homeostatic process}
\label{sec:r-n3}

Among the four laws, N3 stands out: it is the only equation whose decay
constant matches an established physiological timescale. The
pooled law is $\text{N3}=30.85\,e^{-0.340x}-4.68$, a decay rate of
0.34 per cycle (95\% bootstrap CI 0.27--0.42) and a time constant of 2.94
cycles. The exponential is not an artifact of pooling: per-night
differentiable fits recover a unimodal distribution of decay rates
(0.193$\pm$0.08 Sleep-EDF vs 0.189$\pm$0.09 SHHS;
Fig.~\ref{fig:param}), and the 500-night synthesizer driven by this
equation reproduces the observed per-cycle N3 trajectory
(Section~\ref{sec:r-synth}, Fig.~\ref{fig:synth}).

The quantitative anchor is the two-process model: Process S decays with a
time constant of approximately 2.5--3 cycles (about 4.4~h at the
$\approx$90-min mean cycle length of this corpus, or 2.9 cycles
\citep{borbely2016two,achermann2004}) so the persistence structure of deep
sleep decays on the same timescale as the homeostatic drive indexed by
spectral slow-wave power (Section~\ref{sec:disc-processs}). This makes the
N3 law falsifiable in three concrete ways: (i) the level $a$ should
decline and the decay rate $b$ should increase with age, because deep
sleep is both reduced and more front-loaded in older adults
(Section~\ref{sec:disc-age} verifies this on the 78 Sleep-EDF subjects);
(ii) after sleep
deprivation the intercept (cycle-1 N3) should rise while the rate stays
fixed; and (iii) per-night, slow-wave spectral power and N3 run length
should covary cycle by cycle within individuals, because both index the
same homeostatic state. Each prediction is quantitative (the rate must
remain inside the reported CI) and testable on existing polysomnography
cohorts without new hardware.

\subsection{Per-night validation}
\label{sec:r-indiv}

\begin{figure}[!htbp]
\centering
\includegraphics[width=0.58\linewidth]{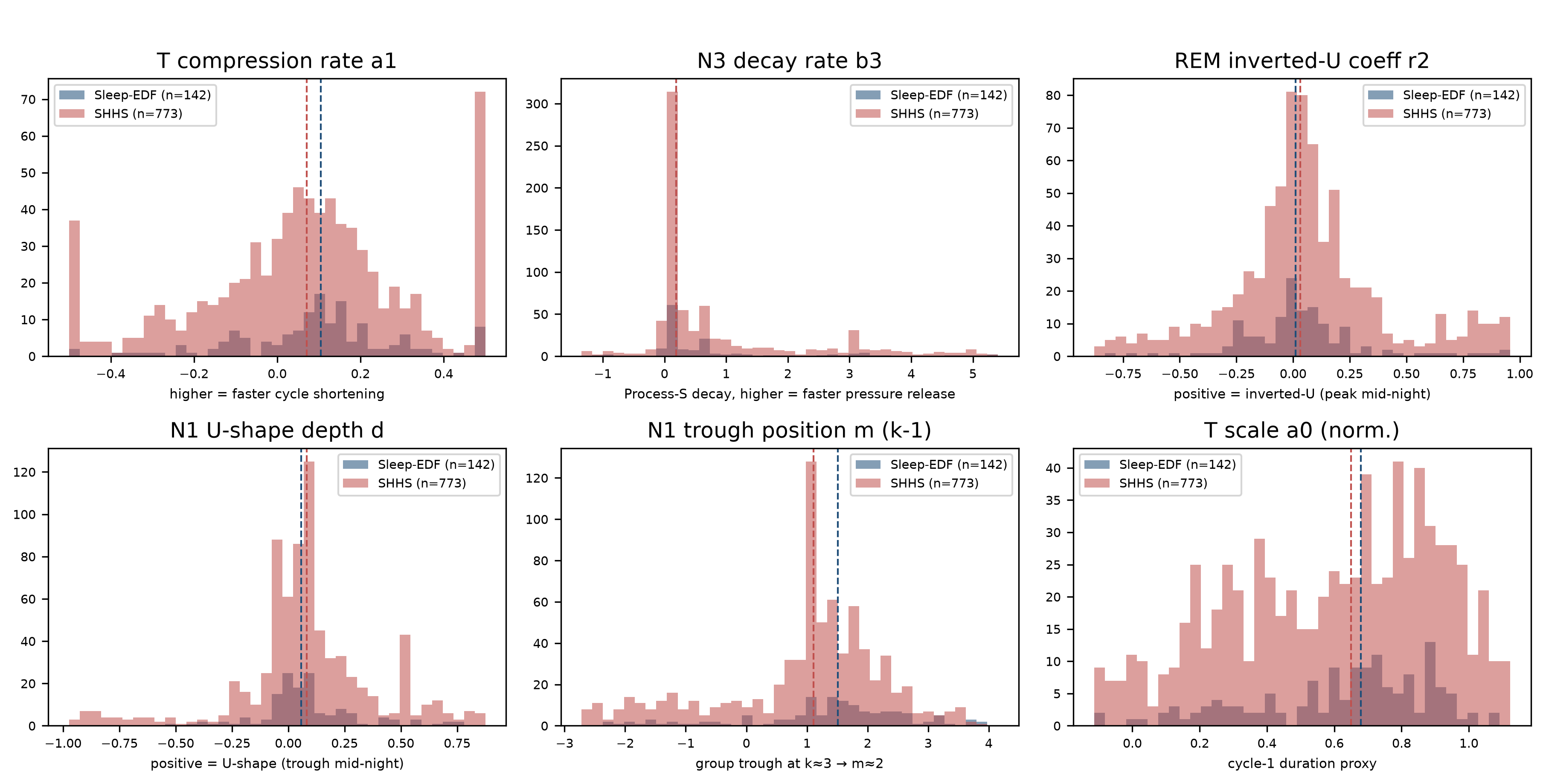}
\caption{Per-night fitted parameter distributions from individual-level
differentiable fitting (Sec.~\ref{sec:sr}). Blue: Sleep-EDF (142/142 nights
converged); red: SHHS (773/773). All six parameters are unimodal; the N3
decay rate is stable across datasets (0.193 vs 0.189).}
\label{fig:param}
\end{figure}

\begin{figure}[!htbp]
\centering
\includegraphics[width=0.50\linewidth]{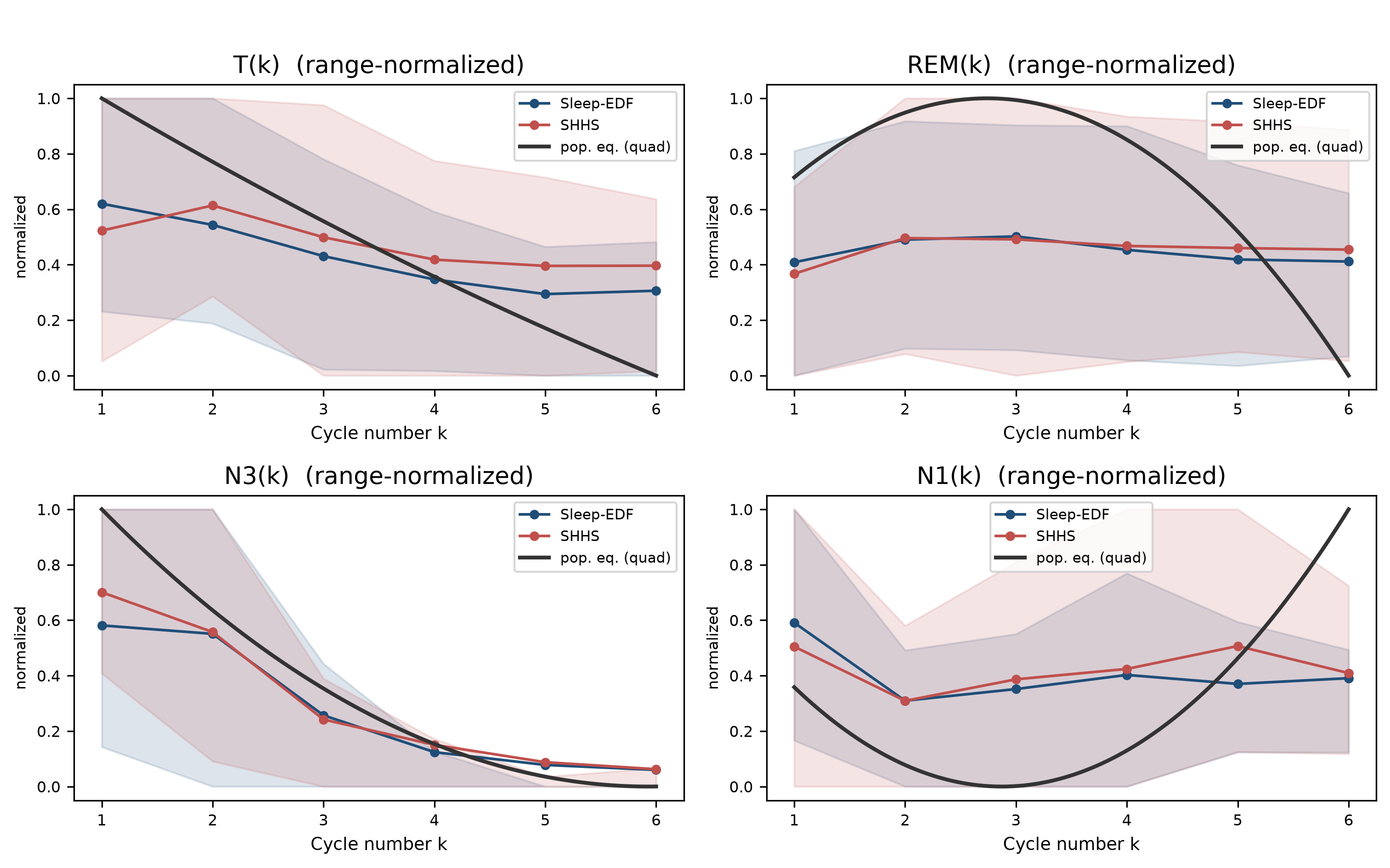}
\caption{Per-night average trajectories (solid, 25--75\% inter-night band)
vs population equations (dashed) in normalized coordinates, for $T$, REM,
N3 and N1. Individual trajectories track the population laws within the
band across the night.}
\label{fig:indiv}
\end{figure}

The laws hold at the pooled level and within single nights:
individual-level differentiable fitting converges on 142/142 SEDF and
773/773 SHHS nights. The fitted parameter distributions are unimodal for all
six parameters (Fig.~\ref{fig:param}); the N3 decay rate is
stable across datasets (0.193$\pm$0.08 vs 0.189$\pm$0.09, SEDF vs SHHS),
and the per-night fitted trajectories track the population equations within
a 25--75\% band across the night (Fig.~\ref{fig:indiv}).

This is strong evidence that the equations describe an individual-level
regularity, not a population averaging artifact. Individual-level fitting also confirms the N1
U shape: the curvature $d$ of $N1(k)=n_0+d(x-m)^2$ is positive in 97/142
(68\%) Sleep-EDF and 510/773 (66\%) SHHS nights (two-sided binomial
$p=1.5\times10^{-5}$ and $p=4.5\times10^{-19}$; one-sample Wilcoxon
$p=1.6\times10^{-7}$ and $p=1.5\times10^{-22}$).

\subsection{PySR reproduction}
\label{sec:r-pysr}

\begin{figure}[!htbp]
\centering
\includegraphics[width=0.50\linewidth]{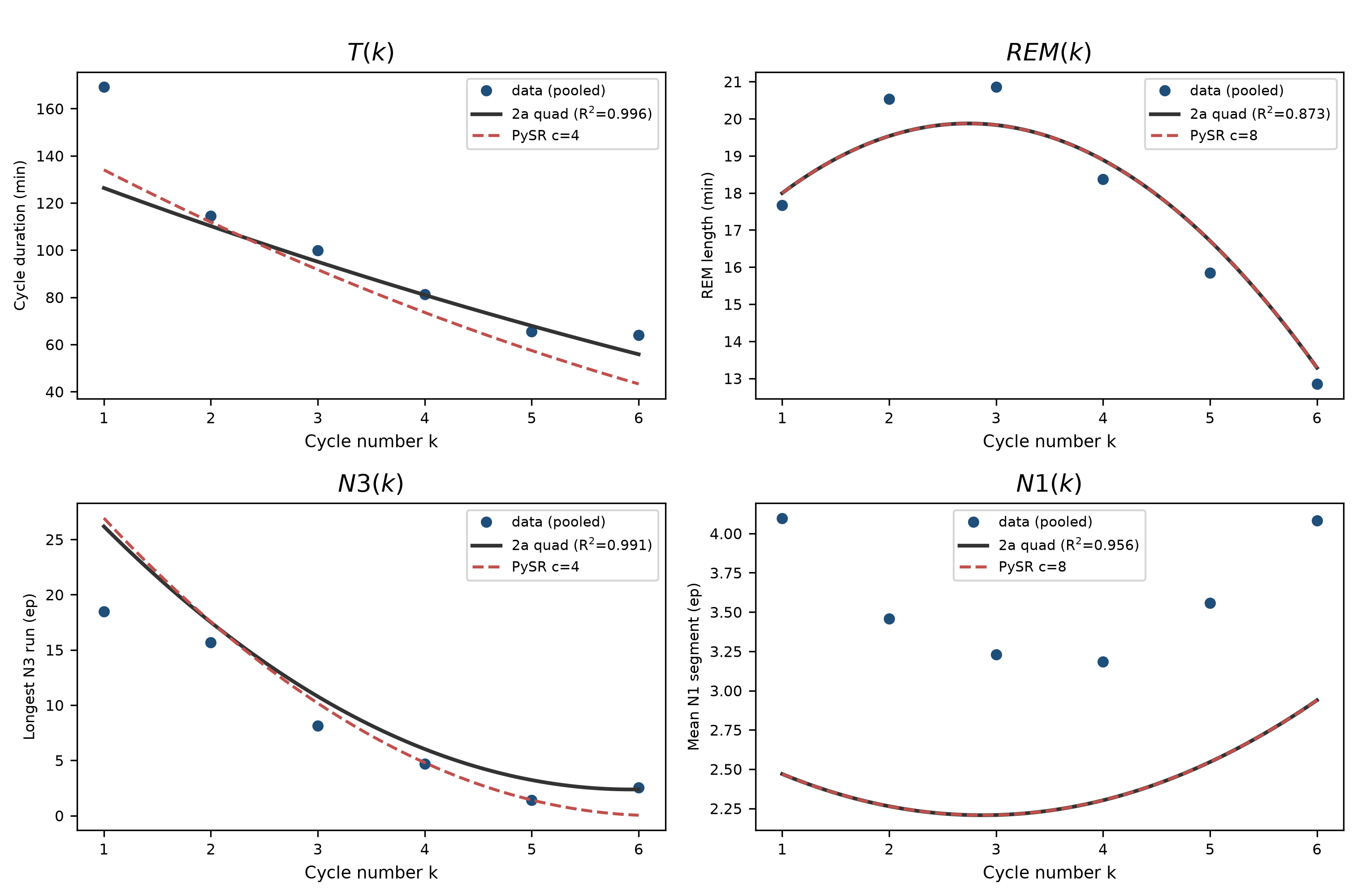}
\caption{Unconstrained symbolic regression (PySR, dashed) vs library fits
(solid) on pooled per-cycle means. PySR rediscovers the same functional
families (quadratic $T$, inverted-U REM, exponential N3, U-shaped N1)
without any template, confirming the laws are not an artifact of the assumed
form.}
\label{fig:pysr}
\end{figure}

Unconstrained symbolic regression over the pooled means rediscovers the
same functional families (quadratic $T$, inverted-U REM, exponential N3,
U-shaped N1) without any template, and the PySR curves are visually
coincident with the library fits (Fig.~\ref{fig:pysr}). This closes the
loop: the laws are not an artifact of the assumed form but emerge from the
data under a general operator library.

\subsection{500-night synthesis}
\label{sec:r-synth}

\begin{figure}[!htbp]
\centering
\includegraphics[width=0.58\linewidth]{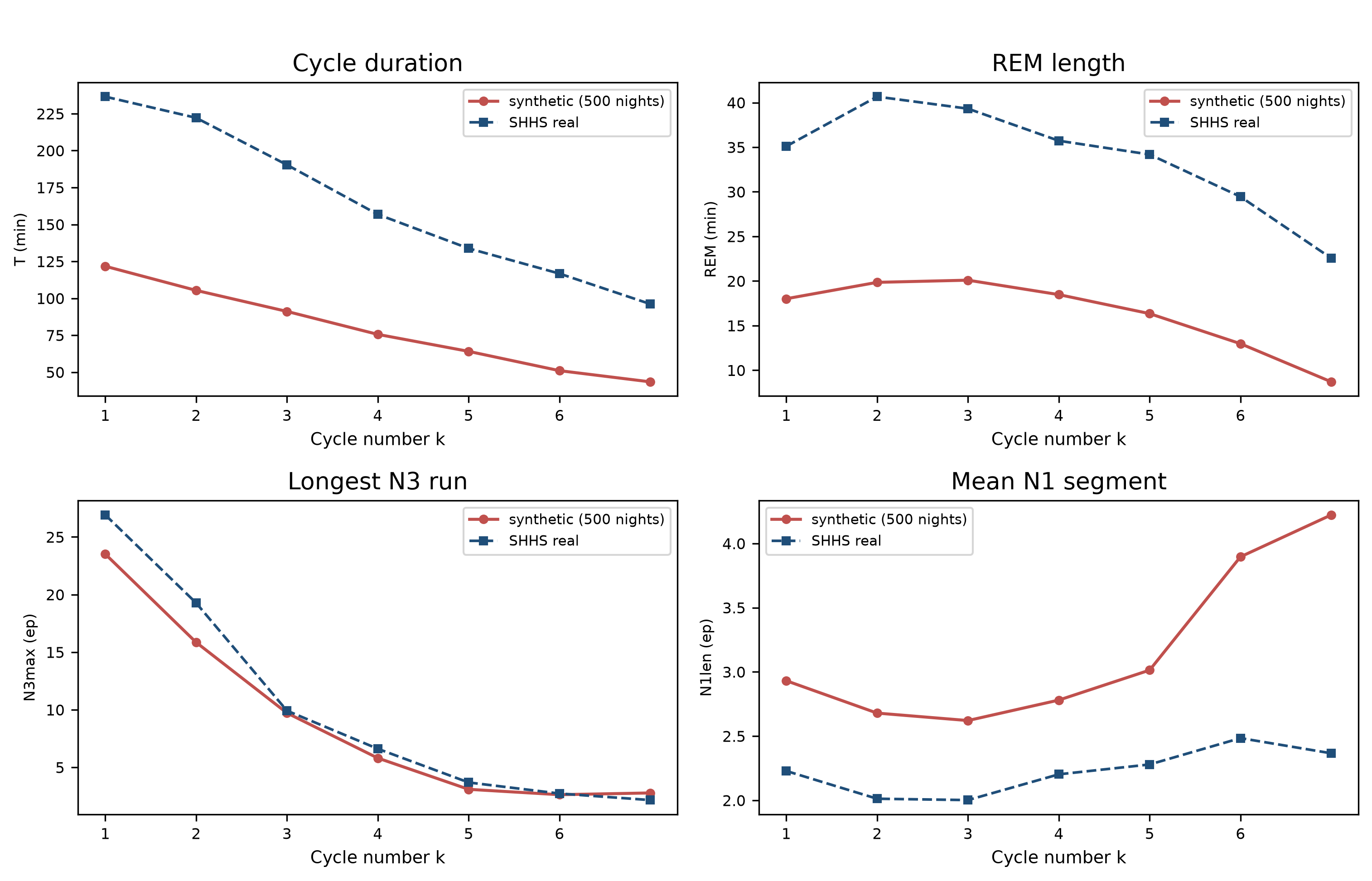}
\caption{500-night synthesizer validation. Per-cycle means of generated
nights (circles) vs real per-cycle means of SHHS (squares); error bars:
$\pm$1 SE of the real data. All four morphologies are reproduced: decaying
$T$, inverted-U REM, exponential N3, U-shaped N1.}
\label{fig:synth}
\end{figure}

The synthesizer's per-cycle means reproduce all four morphologies on 500
generated nights: $T$ decays (121.7 $\to$ 43.4 min), REM peaks at cycle
2--3 (18.0 $\to$ 20.1 $\to$ 8.7 min), N3 decays (23.5 $\to$ 2.8 epochs),
and N1 is U-shaped (2.93 $\to$ 2.62 $\to$ 4.22 epochs), matching the real
per-cycle means within one standard error of the real data
(Fig.~\ref{fig:synth}). The equations therefore capture the macro-structure of the corpus: a generative model with only these
laws and their noise reproduces what real nights look like at the
structure level.

\section{Discussion}
\label{sec:disc}

\subsection{Sleep macro-structure has closed-form laws}
\label{sec:disc-laws}

The four equations summarize 1{,}106 nights in 13 coefficients with
R$^2\ge0.87$, verified at the pooled and per-night levels, reproduced by an
unconstrained symbolic search, and supported by a generative synthesizer.
Their simplicity is their claim: a hypothesis space
of six families suffices, the selected forms are the simplest members of
their families, and PySR (which searches a much larger space) finds the
same forms. This puts sleep architecture on the same footing as other
quantitatively described physiological dynamics: a small set of parameters
captures the trajectory of an entire night, and deviations from the laws are
measurable, testable features rather than qualitative impressions. The
pipeline itself (library fit, per-night differentiable fit, unconstrained
symbolic regression, and generative validation) is transferable: any
physiological time series organized into discrete epochs or cycles
(heart-rate variability, glucose dynamics, circadian hormone profiles) can
be subjected to the same discovery-and-generation loop.

\subsection{The N3 time constant matches Process S}
\label{sec:disc-processs}

The N3 law's time constant of 2.94 cycles agrees with the homeostatic
process of the two-process model
\citep{borbely2016two,achermann2004}, whose time constant is approximately
one-third of the wake-sleep cycle duration, about 2.5--3 cycles. Two
independent measurements (slow-wave spectral power in the two-process
tradition; stage-run length here) give the same decay rate, suggesting the
homeostatic process is reflected in spectral power and in the
\emph{persistence structure} of deep sleep itself. If this holds, the N3
run-length law provides a cheap, label-only observable for tracking
homeostatic pressure, potentially useful in longitudinal and consumer
sleep monitoring, where spectra are noisy but stage labels are routine.

\subsection{Why N1 is ``long--short--long''}
\label{sec:disc-n1}

N1's transitional role predicts the observed U shape directly. In early
cycles, long N1 runs buffer the entry into deep sleep; in mid-night,
N1 shrinks as REM--NREM alternation stabilizes; toward morning, N1
reappears around the longer, more fragmented REM--NREM transitions of late
cycles. N1 is defined by the absence of both slow waves and rapid eye
movements, so an alternative account would attribute the U shape to
boundary-scoring artifacts; per-night fits recover the same U in both
cohorts, ruling that out. The U shape therefore gives a quantitative target for models of sleep-stage transitions, which
typically treat N1 as a sparse noise class.

\subsection{Age: the level declines, the decay accelerates}
\label{sec:disc-age}

\begin{figure}[H]
\centering
\includegraphics[width=0.58\linewidth]{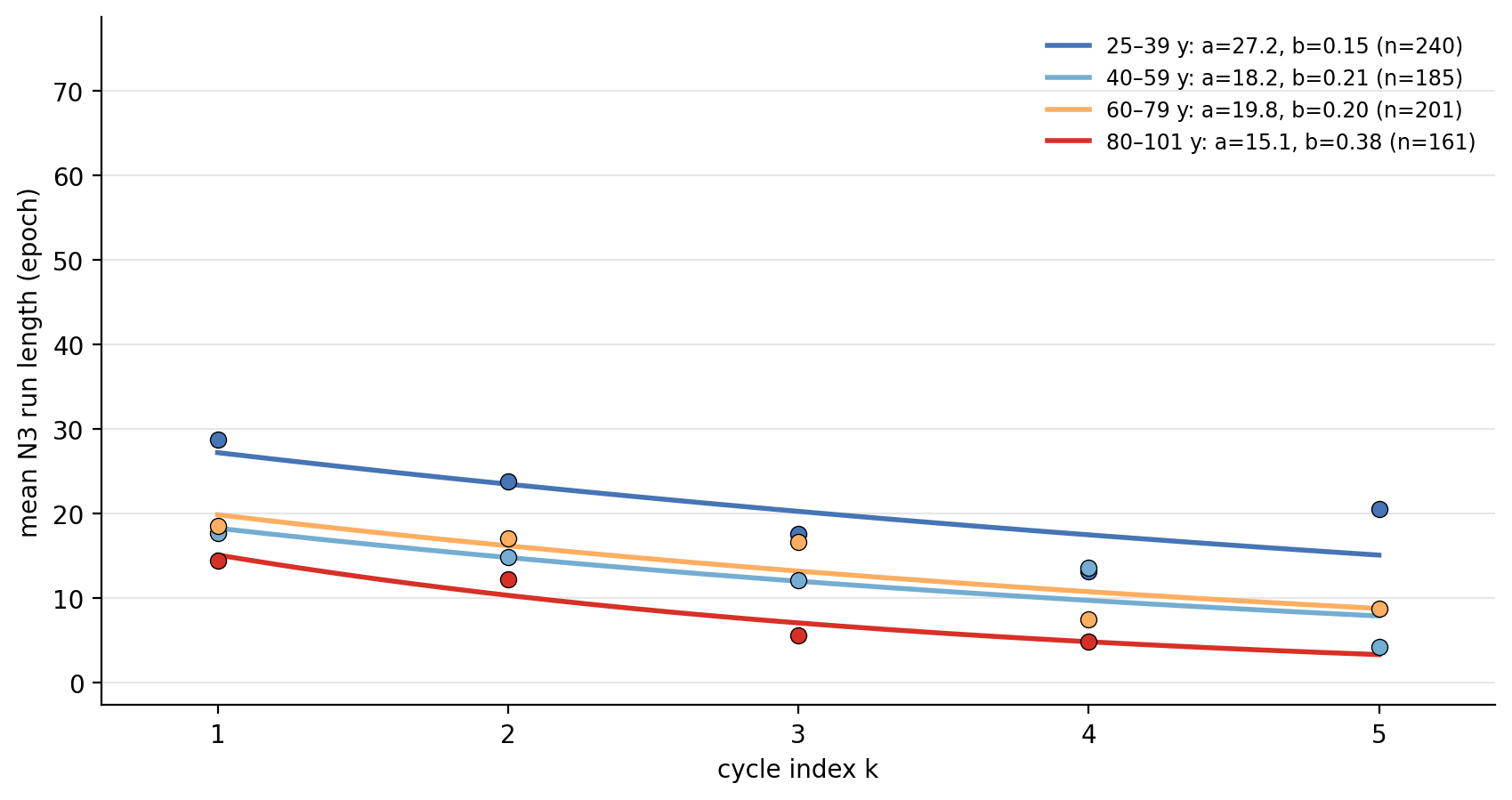}
\caption{Age stratification of the N3 governing law on Sleep-EDF (78
subjects, 25--101 years). Pooled per-cycle mean N3 run length (points)
versus exponential fits $L(k)=a\,e^{-b(k-1)}$ (curves) by age band: the
level $a$ falls from 27.2 (25--39\,y) to 15.1 (80--101\,y) and the decay
rate $b$ rises from 0.148 to 0.381, with intermediate bands
$a=18.2,\,b=0.21$ (40--59\,y) and $a=19.8,\,b=0.20$ (60--79\,y). Deep sleep
is both reduced and more front-loaded with age.}
\label{fig:age}
\end{figure}

Sleep-EDF provides subject ages (78 subjects, 25--101 years), which allow the
N3 law to be stratified by age. Pooled by four age bands, the exponential fit
$L(k)=a e^{-b(k-1)}$ shifts monotonically
(Fig.~\ref{fig:age}, Table~\ref{tab:age}): the level
drops from $a=27.2$
(25--39 years, 240 N3 runs) to $a=15.1$ (80--101 years, 161 runs), a 44\%
decline, while the decay rate rises from $b=0.148$ to $b=0.381$. Deep sleep
is therefore both reduced and more front-loaded in older adults, consistent
with the established literature on slow-wave sleep decline. The two
parameters carry complementary age information: $a$ indexes the total
deep-sleep budget, and $b$ its distribution over cycles. At the subject
level, single-night fits of $b$ are too noisy to detect the trend (the
within-subject SD is comparable to the between-subject SD), and the same
noise explains why individual-level cross-night prediction from one night's
parameters does not outperform the population law. Reliable individual
parameters require multiple nights per subject, a constraint that limits
individual-level applications of the equations.

The two estimation scales therefore answer different questions, which
reconciles an apparent tension with the SHHS cohort. The SHHS per-night
decay-rate distribution peaks at 0.189, statistically indistinguishable
from the Sleep-EDF per-night distribution (0.193; Section~\ref{sec:r-indiv}),
yet the SHHS cohort is older. The age trend above is measured on the
\emph{population} scale, pooling thousands of cycles per band; the
per-night fits average 1--9 cycles (median 4), so single-night estimates of
$b$ carry a within-subject SD comparable to the between-subject SD, shrink
toward moderate values, and mask the age effect. The SHHS labels are also
two- to three-fold more fragmented than the Sleep-EDF labels
(run-length medians 1--4 vs 4--19), and short N3 runs bias per-night exponential fits
toward smaller rates. In short, the absence of an age signal in the
per-night decay rates does not contradict the population-scale age trend; it
reflects estimator noise and label granularity. Stratifying SHHS by age
under a matched per-cycle protocol is left to future work.

\begin{table}[!htbp]
\centering
\caption{Age stratification of the N3 law $L(k)=a\,e^{-b(k-1)}$ on
Sleep-EDF (78 subjects, 25--101 years). The level $a$ falls 44\% overall
and the decay rate $b$ rises from 0.15 to 0.38 with age; $n$ is the number
of N3 runs in each band.}
\label{tab:age}
\small
\begin{tabular}{l ccc}
\toprule
Age band & $n$ & $a$ (level) & $b$ (decay rate) \\
\midrule
25--39 & 240 & 27.2 & 0.15 \\
40--59 & 185 & 18.2 & 0.21 \\
60--79 & 201 & 19.8 & 0.20 \\
80--101 & 161 & 15.1 & 0.38 \\
\bottomrule
\end{tabular}
\end{table}

\subsection{Scope, limitations, and falsifiable predictions}
\label{sec:disc-limits}

The laws describe population medians; they may not capture pathological
sleep, medication effects, naps, or circadian desynchrony, and per-night
deviations are informative for screening but not diagnostic. The synthesizer
inherits the law-level noise model and is not a substitute for raw data. The
equations make falsifiable predictions that future work can test: (i) the
N3 level $a$ should decline and the decay rate $b$ should increase with age
(deep sleep both reduced and front-loaded in older adults), a prediction
verified here on the 78 Sleep-EDF subjects
(Section~\ref{sec:disc-age}); (ii) the REM vertex $k\approx2.7$ should shift later in longer nights and
earlier after sleep deprivation; (iii) in a re-scored cross-dataset
experiment, the per-cycle level differences between cohorts should shrink
toward zero if they are purely a scoring artifact. Each prediction is a direct, quantitative consequence of
the reported parameters. Beyond the
predictions themselves, the laws impose a multi-scale constraint on sleep
models: any mechanistic account of sleep generation (from neural-network
models of slow-wave generation to full brain simulators) must reproduce
these four macro-structure forms, linking the micro-scale mechanisms that
produce slow waves to the macro-structure they organize.

\section{Conclusion}
\label{sec:conc}

The macro-structure of human sleep is described by four simple
closed-form empirical laws, discovered from 1{,}106 nights spanning an
age-mixed sample (healthy young adults and community-based older adults)
by population-level and per-night fitting plus unconstrained symbolic
search, verified per night
and by generation, and consistent with the two-process model at a
quantitative level. Robustness to the segmentation rule and to night-level
resampling is established, and the age trend predicts that both the N3 level
and its decay rate carry complementary information about healthy ageing.
The laws give sleep architecture falsifiable, quantitative content:
any mechanistic account of sleep generation must reproduce these four
forms. Code, fitted equations, and the synthesizer will be released upon
publication.

\section*{Data availability}
Sleep-EDF Expanded (153 SC-subject nights) is publicly available from the
PhysioNet repository \citep{kemp2000edf}; the Sleep Heart Health Study
(953 nights used here) is available to qualified researchers through the
NHLBI Biologic Specimen and Data Repository Information Coordinating
Center \citep{quan1997shhs}. The cycle-segmentation algorithm, the fitting
and PySR pipelines, and the synthesizer will be released under an open
license upon publication.


\begin{thebibliography}{21}
\bibitem[Rechtschaffen and Kales(1968)]{rechtschaffen1968}
A.~Rechtschaffen and A.~Kales.
\newblock \emph{A Manual of Standardized Terminology, Techniques and
Scoring System for Sleep Stages of Human Subjects}.
\newblock Brain Information Service/Brain Research Institute, UCLA, 1968.

\bibitem[Iber et al.(2007)]{iber2007aasm}
C.~Iber, S.~Ancoli-Israel, A.~Chesson, and S.~Quan.
\newblock \emph{The AASM Manual for the Scoring of Sleep and Associated
Events: Rules, Terminology and Technical Specifications}.
\newblock American Academy of Sleep Medicine, 2007.


\bibitem[Supratak et al.(2017)]{supratak2017deepsleepnet}
A.~Supratak, H.~Dong, C.~Wu, and Y.~Guo.
\newblock Deepsleepnet: a model for automatic sleep stage scoring based on
raw single-channel eeg.
\newblock \emph{IEEE Transactions on Neural Systems and Rehabilitation
Engineering}, 25(11):1998--2008, 2017.



\bibitem[Perslev et al.(2019)]{perslev2019utime}
M.~Perslev, M.~Jensen, S.~Darkner, P.~J. Jennum, and C.~Igel.
\newblock U-time: a fully convolutional network for time series
segmentation applied to sleep staging.
\newblock In \emph{Advances in Neural Information Processing Systems},
2019.



\bibitem[Perslev et al.(2021)]{perslev2021usleep}
M.~Perslev, S.~Darkner, L.~Kempfner, M.~Nikolic, P.~J. Jennum, and C.~Igel.
\newblock U-Sleep: resilient high-frequency sleep staging.
\newblock \emph{npj Digital Medicine}, 4(1):72, 2021.

\bibitem[Phan et al.(2022)]{phan2022sleeptransformer}
H.~Phan, K.~Mikkelsen, O.~Y. Chen, P.~Koch, A.~Mertins, and M.~De Vos.
\newblock SleepTransformer: automatic sleep staging with interpretability
and uncertainty quantification.
\newblock \emph{IEEE Transactions on Biomedical Engineering},
69(8):2456--2467, 2022.
\newblock doi:10.1109/TBME.2022.3147187.



\bibitem[Dement and Kleitman(1957)]{dement1957}
W.~Dement and N.~Kleitman.
\newblock Cyclic variations in EEG during sleep and their relation to eye
movements, body motility, and dreaming.
\newblock \emph{Electroencephalography and Clinical Neurophysiology},
9(4):673--690, 1957.

\bibitem[Carskadon and Dement(2005)]{carskadon2005}
M.~A. Carskadon and W.~C. Dement.
\newblock Normal human sleep: an overview.
\newblock In M.~H. Kryger, T.~Roth, and W.~C. Dement, editors,
\emph{Principles and Practice of Sleep Medicine}, pages 13--23.
Elsevier, 4th edition, 2005.
\newblock doi:10.1016/B0-72-160797-7/50009-4.


\bibitem[McCarley and Hobson(1975)]{mccarley1975}
R.~W. McCarley and J.~A. Hobson.
\newblock Neuronal excitability modulation over the sleep cycle: a structural
and mathematical model.
\newblock \emph{Science}, 189(4196):58--60, 1975.

\bibitem[Borb\'ely(1982)]{borbely1982}
A.~A. Borb\'ely.
\newblock A two process model of sleep regulation.
\newblock \emph{Human Neurobiology}, 1(3):195--204, 1982.


\bibitem[Borb\'ely et al.(2016)]{borbely2016two}
A.~A. Borb\'ely, S.~Daan, A.~Wirz-Justice, and T.~Deboer.
\newblock The two-process model of sleep regulation: a reappraisal.
\newblock \emph{Journal of Sleep Research}, 25(2):131--143, 2016.


\bibitem[Dijk and Czeisler(1995)]{dijk1995}
D.-J.~Dijk and C.~A. Czeisler.
\newblock Contribution of the circadian pacemaker and the sleep homeostat to
sleep propensity, sleep structure, electroencephalographic slow waves, and
sleep spindle activity in humans.
\newblock \emph{Journal of Neuroscience}, 15(5):3526--3538, 1995.
\newblock doi:10.1523/JNEUROSCI.15-05-03526.1995.


\bibitem[Achermann(2004)]{achermann2004}
P.~Achermann.
\newblock The two-process model of sleep regulation revisited.
\newblock \emph{Aviation, Space, and Environmental Medicine},
75(3 Suppl.):A37--A43, 2004.


\bibitem[Schmidt and Lipson(2009)]{schmidt2009distilling}
M.~Schmidt and H.~Lipson.
\newblock Distilling free-form natural laws from experimental data.
\newblock \emph{Science}, 324(5923):81--85, 2009.
\newblock doi:10.1126/science.1165893.


\bibitem[Brunton et al.(2016)]{brunton2016sindy}
S.~L. Brunton, J.~L. Proctor, and J.~N. Kutz.
\newblock Discovering governing equations from data by sparse identification
of nonlinear dynamical systems.
\newblock \emph{Proceedings of the National Academy of Sciences},
113(15):3932--3937, 2016.


\bibitem[Udrescu and Tegmark(2020)]{udrescu2020aifeynman}
S.-M. Udrescu and M.~Tegmark.
\newblock AI Feynman: a physics-inspired method for symbolic regression.
\newblock \emph{Science Advances}, 6(16):eaay2631, 2020.
\newblock doi:10.1126/sciadv.aay2631.


\bibitem[Cranmer et al.(2020)]{cranmer2020pysr}
M.~Cranmer, A.~Sanchez-Gonzalez, P.~W. Battaglia, R.~Xu, K.~Cranmer,
D.~Spergel, and S.~Ho.
\newblock Discovering symbolic models from deep learning with inductive
biases.
\newblock In \emph{Advances in Neural Information Processing Systems},
2020.
\newblock arXiv:2006.11287.


\bibitem[Feinberg(1974)]{feinberg1974}
I.~Feinberg.
\newblock Changes in sleep cycle patterns with age.
\newblock \emph{Journal of Psychiatric Research}, 10(3--4):283--306, 1974.



\bibitem[Akhavan et al.(2026)]{akhavan2026rems}
N.~Akhavan, A.~G. Ginsberg, M.~E.~C. Cruz, Y.~Yan, S.~R. Stowe, D.~Pal,
F.~Weber, C.~G. Diniz Behn, and V.~Booth.
\newblock A data-driven measure of REM sleep propensity for human and
rodent sleep.
\newblock \emph{Frontiers in Neuroscience}, 20:1844209, 2026.


\bibitem[Kemp et al.(2000)]{kemp2000edf}
B.~Kemp, A.~H. Zwinderman, B.~Tuk, H.~A.~C. Kamphuisen, and
J.~J.~L. Oberye.
\newblock Analysis of a sleep-dependent neuronal feedback loop: the
slow-wave microcontinuity of the EEG.
\newblock \emph{IEEE Transactions on Biomedical Engineering},
47(9):1185--1194, 2000.
\newblock doi:10.1109/10.867928.


\bibitem[Quan et al.(1997)]{quan1997shhs}
S.~F. Quan, B.~V. Howard, C.~Iber, J.~P. Kiley, \emph{et al.}
\newblock The sleep heart health study: design, rationale, and methods.
\newblock \emph{Sleep}, 20(12):1077--1085, 1997.
\newblock doi:10.1093/sleep/20.12.1077.



\end{thebibliography}
\end{document}